\documentclass[preprint,12pt]{elsarticle}
\usepackage{graphicx}
\usepackage{amsmath}
\usepackage{booktabs}
\usepackage{makecell}
\usepackage{float}
\usepackage{siunitx}
\usepackage{hyperref}
\journal{Computerized Medical Imaging and Graphics}
\begin{document}
\begin{frontmatter}
\title{Dataset-Origin Signatures and Shortcut Learning in Screening Mammography AI: A Cross-Dataset Case Study}
\author[1]{Parham Hajishafiezahramini}
\author[1]{Matthew Hamilton}
\author[1]{Oscar Meruvia Pastor}
\author[2]{Edward Kendall}
\affiliation[1]{organization={Department of Computer Science, Memorial University of Newfoundland},
                country={Canada}}
\affiliation[2]{organization={Division of Biomedical Sciences, Faculty of Medicine, Memorial University of Newfoundland},
                country={Canada}}


\begin{abstract}

Reliable AI for screening mammography depends on training data that reflects the low cancer prevalence and subtle presentation of a genuine screening population. It is often assumed that supplementing such data with biopsy-confirmed cases from external datasets will improve performance. We test this assumption using the Newfoundland and Labrador Breast Screening Dataset (NLBSD) alongside two abnormal-enriched public datasets, CBIS-DDSM and CMMD.
Using an EfficientNet-B5 encoder with Mammo-CLIP mammography-pretrained weights as a frozen linear probe, under consistent preprocessing and patient-level splits, we first established reference performance on NLBSD (AUC-ROC = 0.737, 95\% CI [0.686, 0.785]). We then augmented the training set with external positive cases while holding the NLBSD test set fixed. Performance declined in every configuration (AUC-ROC = 0.620--0.644; DeLong test, Holm-corrected $p < 0.05$), and the degradation grew monotonically as further external sources were added. Domain-matched evaluation yielded modest gains only where training and test domains coincided, and none surpassed the NLBSD-only model.

To probe the cause, we ran a complementary diagnostic that reframed the task as identifying each exam's dataset of origin rather than its clinical label. Despite identical preprocessing and normalization, the datasets were separated almost perfectly, indicating that dataset-specific characteristics dominate the learned representation even within a mammography-pretrained encoder.
Taken together, these findings show that naïvely mixing abnormal-only datasets introduces domain shift that outweighs the value of the added positive cases, even under identical preprocessing. Differences in acquisition hardware, intensity mapping, and dataset construction survive normalization and erode performance. This helps explain why the many available mammography datasets remain effectively siloed rather than functioning as one combined resource.

Our aim is not to present a state-of-the-art model but to characterize why heterogeneous screening datasets resist naïve pooling, and to argue for domain-aware strategies before any such combination is attempted.
\end{abstract}
\end{frontmatter}
\section{Introduction}

Breast cancer is one of the most common malignancies affecting women globally \cite{Sung2021-hq}. Its prevalence has now surpassed that of lung cancer, making it the most frequently identified cancer worldwide \cite{Arnold2022-vq}. In 2020 alone, an estimated 2.3 million new cases were reported \cite{Sung2021-hq}. One in eight women in the United States will develop invasive breast cancer during their lifetime \cite{Plunkett2022-us,Parada2019-ww}, and in Canada, survival outcomes range from 99\% for early-stage tumors to 30\% for later-stage disease \cite{ccs2023breastcancer}.

Early detection through symptom awareness, widespread mammography screening, and improvements in treatment have substantially reduced breast cancer mortality \cite{Tabar2022-pv, Tabar2019-wr, Duffy2020-we,Sardanelli2017-ua}. Screening programs use mammography to identify early abnormalities \cite{Sweeney2017-ow}, and typically begin when women are in their forties, when the expected population prevalence is only about 1\% \cite{Kendall2025-mv,Giaquinto2024-yk}. This low prevalence is precisely what makes screening data difficult to model: genuinely suspicious findings are rare, subtle, and vastly outnumbered by normal exams. Yet approximately 1 in 8 cancers are still missed during screening \cite{acs_mammograms}, underscoring the need for models that reflect the true distribution of a screening population rather than artificially enriched case mixes.

Despite these motivations, relatively few studies address automated detection methods that distinguish normal from abnormal screening cases~\cite{Jafari2023-zu, Kendall2013-ec, Kendall2014-zg, Abdikenov2025-jd}. A central reason is the limited availability of biopsy-confirmed datasets that reflect screening workflows, particularly for separating normal from suspicious cases. Publicly accessible datasets such as CBIS-DDSM \cite{Lee2017-fa} and CMMD \cite{Cai2023-wk} do contain biopsy-confirmed cancers, but they do not resemble a true screening population: they consist mostly of clear malignant cases and a small number of biopsy-proven benign lesions, with almost no genuinely normal or subtle findings. A further reason for the field's emphasis on benign–malignant classification is simply that such datasets are more class-balanced, and therefore easier to train and evaluate on.

In other machine-learning domains, such as text classification, increasing minority-class samples or applying oversampling can improve model discrimination \cite{Taskiran2025-rt}. Whether this holds in mammography is contested. Some studies argue that larger and more diverse multi-site training data improve generalization across populations and scanners \cite{Velarde2024-cm}, while others show that strong performance on one dataset frequently fails to transfer to external datasets, and that combining multiple sources does not reliably recover this loss \cite{Wang2020-ez}. Critically, reported gains from additional data typically arise only when domain shift is explicitly addressed through augmentation or harmonization, rather than from naïve pooling. Whether augmenting a real screening cohort with enriched positive datasets improves or degrades screening performance therefore remains an open question.

In this study, we intentionally begin with a simple, controlled experimental setup. As a feature extractor, we use an EfficientNet-B5 encoder initialized with Mammo-CLIP~\cite{Ghosh2024-lb} mammography-pretrained weights and kept frozen, training only a linear classification head on top (a linear-probe setup). We first train on the Newfoundland and Labrador Breast Screening Dataset (NLBSD) \cite{Kendall2025-jn} alone and evaluate its performance, then examine whether adding external cancer cases from CMMD \cite{Cai2023-wk} and CBIS-DDSM \cite{Lee2017-fa} to the training set improves performance on the fixed NLBSD test set. Throughout, all datasets pass through an identical preprocessing pipeline, including per-image intensity normalization to a fixed range, so that any performance change reflects genuine domain differences rather than superficial discrepancies in image formatting.

Through this design we examine two common assumptions about how additional data and technical variation influence screening-AI performance. The first is that increasing the number of positive cases should inherently strengthen a model's ability to detect abnormalities. The second is that identical preprocessing and per-image intensity normalization are sufficient to neutralize technical differences between datasets, such as acquisition hardware, pixel size, bit depth, and DICOM intensity mapping, so that images from different sources can be treated as a single distribution \cite{Guan2022-bp}.

We find that neither assumption holds: adding external positives degrades performance, and dataset-specific signatures persist after normalization strongly enough that a model can identify an image's dataset of origin almost perfectly. We further assess how well models trained in one setting transfer to another across domain boundaries. Our goal is practical: to determine what genuinely helps a screening-style classifier under conditions resembling day-to-day clinical practice, and what does not.

\subsection{Related Work}

A major barrier in developing AI for breast cancer screening is the lack of datasets that capture the diversity and prevalence patterns of real screening populations. Historically, much of the foundational work in mammographic AI relied on the MIAS and DDSM datasets \cite{Suckling2015-al, heath2001digital}, which, despite their historical value, contain limited case diversity and reflect older imaging technologies. More recent large-scale datasets such as OPTIMAM \cite{Halling-Brown2021-yp}, CSAW \cite{Dembrower2020-ul}, and ADMANI \cite{Frazer2023-zn} provide substantially broader coverage but are not publicly available. Among openly accessible datasets, only NLBSD \cite{Kendall2025-mv} represents a true screening cohort with a normal--suspicious labeling scheme.
\begin{table}[H]
\centering
\scriptsize
\caption{Overview of available breast cancer screening datasets, including year of release, image source, size, accessibility, breast screening program, and dataset type.}
\label{tab:datasets}
\resizebox{\textwidth}{!}{%
\begin{tabular}{l c c c c c c}
\toprule
\textbf{Name} & \textbf{Year} & \textbf{Image Source} & \textbf{No. of Cases} & \textbf{Accessibility} & \makecell{\textbf{Breast Screening}\\\textbf{Program}} & \textbf{Dataset Type} \\
\midrule
DDSM \cite{heath2001digital} & 1996 & Film & 2{,}620 & Public & No & Malignant--Benign \\
CBIS-DDSM \cite{Lee2017-fa} & 2017 & Scanned-Film & 1{,}644 & Public & No & Malignant--Benign \\
MIAS/Mini MIAS \cite{Suckling2015-al} & 1994 & Film & 322 & Public & No & Malignant--Benign \\
INbreast \cite{Moreira2012-si} & 2011 & Digital & 115 & Public & No & Normal--Suspicious \\
OPTIMAM \cite{Halling-Brown2021-yp} & 2020 & Digital & 172{,}282 & Commercial & No & Malignant--Benign \\
CMMD \cite{Cai2023-wk} & 2022 & Digital & 1{,}775 & Public & No & Malignant--Benign \\
VinDr \cite{Nguyen2023-zg} & 2023 & Digital & 5{,}000 & Public & No & Malignant--Benign \\
RSNA \cite{carr2022rsna} & 2022 & Digital & 8{,}000 & Public & No & Normal--Suspicious \\
Emory Breast \cite{Jeong2023-cl} & 2023 & Digital & 116{,}000 & Restricted & No & Malignant--Benign \\
ADMANI \cite{Frazer2023-zn} & 2023 & Digital & 629{,}863 & Restricted & No & Malignant--Benign \\
CSAW \cite{Dembrower2020-ul} & 2023 & Digital & 500{,}000 & Restricted & No & Malignant--Benign \\
NLBSD \cite{Kendall2025-mv} & 2025 & Digital & 5{,}997 & Public & Yes & Normal--Suspicious \\
\specialrule{1pt}{0pt}{0pt}
\end{tabular}
}
\end{table}

While a limited number of studies formulate the task as positive–negative or abnormal–normal classification, these approaches are typically evaluated on artificially balanced or enriched datasets and remain focused on lesion-centric or diagnostic settings rather than population-level screening. Jafari and Karami \cite{Jafari2023-zu} trained CNN-based classifiers on a balanced version of the RSNA dataset and reported accuracy and sensitivity above 92\%. Sahu et al. \cite{Sahu2023-jj} developed a hybrid CNN architecture evaluated on balanced subsets of DDSM and ultrasound datasets, achieving accuracies between 94\% and 98\%. Similarly, Raaj \cite{Raaj2023-li} expanded the MIAS and DDSM datasets through data augmentation and obtained sensitivity and specificity near 98\%.

More recently, Abdikenov et al. \cite{Abdikenov2025-jd} evaluated convolutional and transformer-based multi-view models across several datasets, including the newly released NLBSD. While high performance was achieved on smaller curated datasets (AUC = 0.905 on INbreast), performance declined markedly on NLBSD (maximum AUC = 0.6209), underscoring the difficulty of generalizing to real screening populations. Notably, their models were trained without mammography-specific pretraining; as we show, initializing from a mammography-pretrained encoder yields substantially stronger baseline performance on the same cohort, which we adopt as the starting point for our analysis.

Across these studies, two recurring limitations are evident. First, in real screening workflows, both malignant cancers and biopsy-proven benign abnormalities originate from exams initially classified as “abnormal.” However, available benign–malignant datasets are built from pathology-confirmed cohorts rather than screening labels, and therefore do not reflect the normal–suspicious distinction used in population screening. This construction leads to far higher proportions of abnormal cases and often balanced class ratios, in contrast to actual screening programs where cancer prevalence is below 2\% \cite{Kendall2025-mv, Giaquinto2024-yk}. Such discrepancies raise questions about whether models trained on benign-malignant datasets are suitable for use to train screening program models.

Second, relatively few studies evaluate how models behave when transferred across datasets or when datasets with different characteristics are combined. With substantial variation in imaging equipment, acquisition protocols, population prevalence, and pixel-processing pipelines across institutions, it is not clear a priori whether incorporating external, cancer-enriched datasets will strengthen or degrade performance on an actual screening cohort. Understanding this interaction between dataset composition and model behavior is essential for determining whether AI systems trained in one setting can reliably generalize to another and whether data normalization can be a solution to bridge the gap between acquisition domains.

These open questions motivate the present study. We evaluate both cross-dataset generalization and the impact of mixed-dataset training, focusing on whether augmenting a screening dataset with external cancer-enriched collections improves or degrades performance. By testing models trained on NLBSD alone, CMMD/CBIS-DDSM alone, and various combinations, we aim to characterize how dataset composition and domain differences influence model behavior in a setting that reflects day-to-day clinical screening.

\section{Materials and Methods}

\subsection{Dataset}

\subsubsection{NLBSD}
We employed the Newfoundland and Labrador Breast Screening Program (NLBSD) dataset \cite{Kendall2025-jn, Kendall2025-mv}, which contains 5{,}997 screening cases with CC and MLO views for both breasts. The prevalence of recall-suspicious findings is 1.29\% (149/5{,}997), reflecting real-world screening conditions. Images appear in two native resolutions (2395$\times$3063 and 1915$\times$2295) and are stored in 16-bit grayscale. NLBSD is used as the primary evaluation dataset because it closely approximates a true population-based screening cohort and represents data collected over a continuous time period within a provincial screening program.

\subsubsection{CBIS-DDSM}
CBIS-DDSM provides 1,644 cases, comprising 753 calcification cases and 891 mass cases, each with biopsy-confirmed benign or malignant outcomes. The dataset is a curated subset of the original DDSM collection and consists of screen-film mammograms that were subsequently digitized and standardized for research use, rather than images acquired with native digital mammography systems. As an abnormal-enriched dataset with no normal screening exams, it does not reflect screening prevalence. In this study, it is used only to supply additional abnormal examples for training NLBSD-based models, allowing us to examine the effect of adding positive-heavy external data.

\subsubsection{CMMD}
The Chinese Mammography Database (CMMD) \cite{Cai2023-wk} contains 3{,}712 natively digital mammographic images from 1{,}775 patients, provided in DICOM format, and is divided into two subsets: CMMD1 (1{,}026 biopsy-confirmed benign or malignant cases, 2{,}214 images) and CMMD2 (749 patients, 1{,}498 images, with recorded molecular subtypes). Like CBIS-DDSM, CMMD is abnormal-enriched and pathology-confirmed, with no normal screening exams. It is therefore used only to provide additional abnormal samples during training rather than to characterize screening performance.

\subsection{Data Preprocessing}
All DICOM images were converted to 16-bit PNG using an identical pipeline across every dataset, so that pixel intensities are processed consistently and cross-dataset comparison remains valid. For each image we applied the modality LUT (rescale slope/intercept) followed by the VOI LUT to obtain vendor-consistent presentation values, and inverted \texttt{MONOCHROME1} images so that breast tissue is always bright. Intensities were then normalized per image to the $[0,1]$ range using its own minimum and maximum, without any global or cross-dataset intensity mapping.

To isolate the breast and remove background, labels, and markers, we estimated the background level from the image border and applied a low threshold just above it, retaining all breast tissue including dim fatty margins. We kept the largest connected component, filled enclosed holes by flood-filling the background, and applied a light morphological closing to smooth the boundary. The resulting mask was used to crop to the breast bounding box with a fixed 10\% margin in all directions, and background pixels outside the mask were set to zero. To standardize orientation, all right-breast images were horizontally flipped (using \texttt{numpy}) so that every image is in a consistent left-oriented configuration. The same pipeline was applied identically to NLBSD, CBIS-DDSM, and CMMD, with no dataset-specific tuning or harmonization.

\subsection{Data Augmentation}
To improve robustness to acquisition variability, we applied light geometric and photometric augmentations to the training set. Each training image underwent a random affine transformation combining rotation ($\pm10^\circ$), isotropic scaling ($0.9$--$1.1$), and translation (up to $5\%$ of the image dimensions in each direction), followed by small random brightness ($\pm0.05$) and contrast ($0.90$--$1.10$) perturbations. Horizontal flipping was deliberately not applied, since all images had already been standardized to a single (left) laterality during preprocessing. Augmentations were sampled stochastically at each training iteration and applied only to the training split; validation and test images received no augmentation beyond the shared preprocessing pipeline.

\subsection{Model Setup and Training}
\begin{figure}[htbp]
\centering
\includegraphics[width=0.89\linewidth]{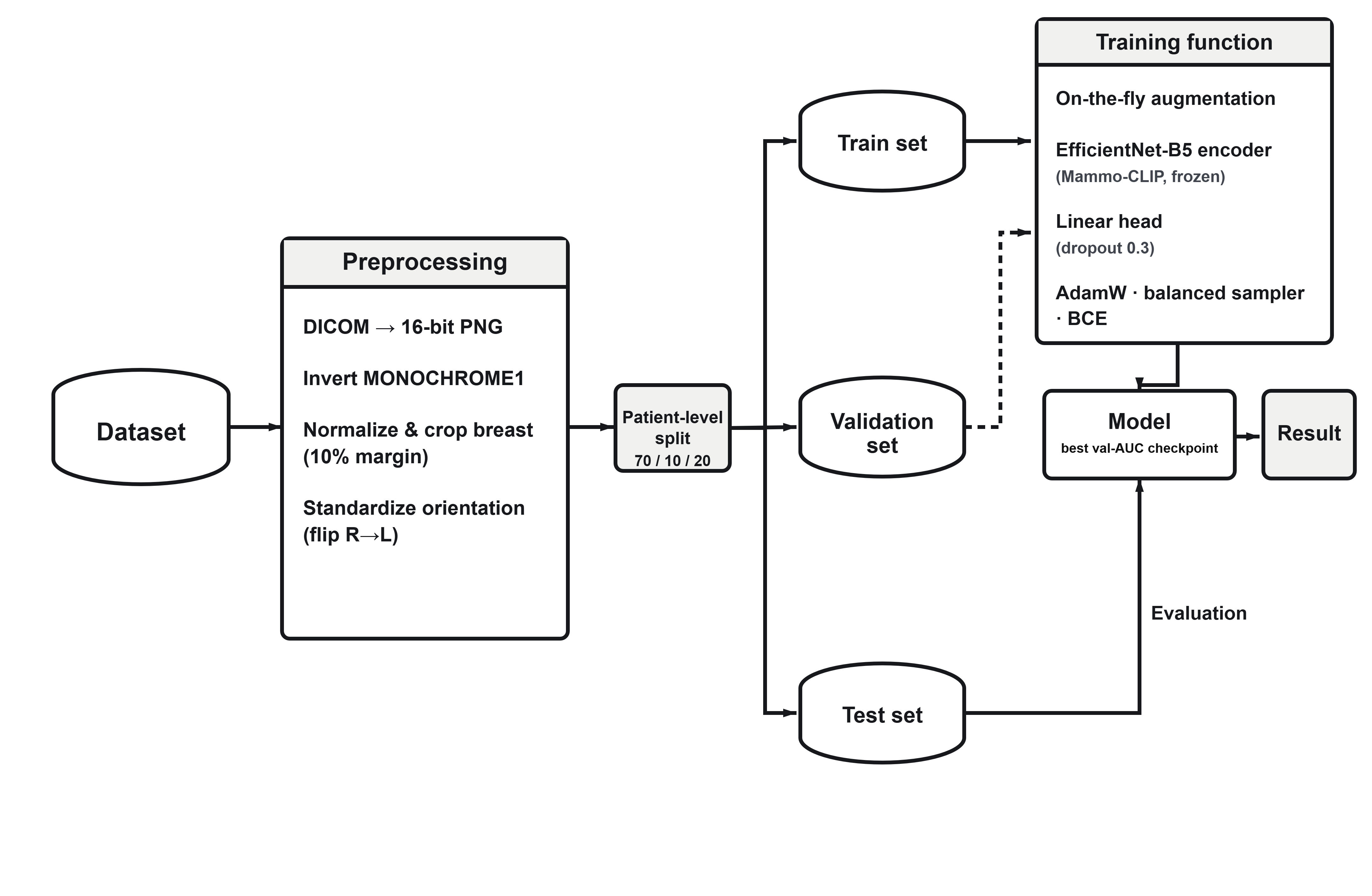}
\caption{Overview of the experimental pipeline, including data preprocessing, patient-level dataset splitting, model configuration, training with data augmentation, and final evaluation on a held-out test set.}
\label{fig:model_setup}
\end{figure}

All experiments were conducted on an IBM high-performance computing cluster, with data loading handled by a multi-threaded pipeline (8 workers per job) to minimize GPU idle time. As the feature extractor, we used an EfficientNet-B5 encoder \cite{pmlr-v97-tan19a} initialized with Mammo-CLIP mammography-pretrained weights~\cite{Ghosh2024-lb} and kept frozen; only a linear classification head (layer normalization, dropout of 0.3, and a single linear layer) was trained on top of the pooled encoder features. Following preprocessing, the data were split at the patient level into 70\% for training, 10\% for validation, and 20\% for testing, so that all images from a given patient appeared in only one subset. Model optimization was performed on the training subset, while the validation subset guided early stopping and operating-threshold calibration, and final performance was assessed exclusively on the held-out test subset. \textbf{Figure~\ref{fig:model_setup}} illustrates the overall workflow.

The linear head was optimized using AdamW~\cite{Loshchilov2017-un} (learning rate $3\times10^{-5}$, weight decay $10^{-4}$) with a batch size of 80 and automatic mixed precision. To counter the severe class imbalance, each mini-batch was constructed with a fixed number of positive cases (20 positives per batch) using a balanced batch sampler, rather than reweighting the loss. The model was trained with a standard binary cross-entropy objective. Training ran for a maximum of 30 epochs with early stopping on validation AUC-ROC (patience of 10 epochs), and the checkpoint with the highest validation AUC was retained for evaluation.

\subsection{Additional Experiments: Dataset Mixing and Generalization}
To examine whether adding abnormal-enriched external datasets with different imaging characteristics improves performance on a true screening population, we conducted a series of controlled experiments using different combinations of NLBSD, CBIS-DDSM, and CMMD during training. In all experiments, negative cases were drawn exclusively from NLBSD, while external datasets supplied additional biopsy-confirmed abnormal cases. All preprocessing and training procedures (including on-the-fly augmentations applied \emph{only} to the training set) were kept identical to the baseline to ensure comparability.

\paragraph{Phase A: Effect of adding external positives (fixed NLBSD test set).}
To isolate the impact of external positive cases, we preserved the original NLBSD test set (20\%, patient-level split) and retrained the model under three conditions (the $^{+}$ superscript denotes that only positive/abnormal cases from the external dataset were added to the NLBSD training set):
\begin{enumerate}
    \item \textbf{NLBSD + CMMD$^{+}$}: NLBSD training set with CMMD positive cases added.
    \item \textbf{NLBSD + CBIS-DDSM$^{+}$}: NLBSD training set with CBIS-DDSM positive cases added.
    \item \textbf{NLBSD + CBIS-DDSM$^{+}$ + CMMD$^{+}$}: NLBSD training set with positives from both external datasets.
\end{enumerate}
Because the test set remains the fixed NLBSD split in all three conditions, this setup directly tests whether increasing the number of abnormal training samples improves performance on a true screening test distribution.

\paragraph{Phase B: Evaluation on mixed-domain test sets.}
To assess robustness in settings where future screening data may include images acquired under different conditions, we repeated the training configurations above and evaluated performance on mixed-domain test sets. In each case, the negative class consisted exclusively of NLBSD screening exams, while the positive class included abnormal cases from NLBSD and, where applicable, from the external datasets. All external cases were partitioned at the patient level, so that no patient contributing positives to a test set appeared in the corresponding training data. Specifically, we considered three mixed-domain test sets:
\begin{enumerate}
    \item \textbf{NLBSD $\cup$ CMMD}: negatives from NLBSD, positives from NLBSD and CMMD;
    \item \textbf{NLBSD $\cup$ CBIS-DDSM}: negatives from NLBSD, positives from NLBSD and CBIS-DDSM;
    \item \textbf{NLBSD $\cup$ CBIS-DDSM $\cup$ CMMD}: negatives from NLBSD, positives from NLBSD, CBIS-DDSM, and CMMD.
\end{enumerate}
These tests reflect performance in a domain-mixed scenario with screening-derived negative cases and are intended solely as a supplementary analysis, not as a substitute for the primary screening evaluation. We did not perform cross-dataset testing on only CBIS-DDSM or CMMD, as these datasets contain no normal screening cases and therefore cannot support a full evaluation (e.g., ROC analysis or specificity computation).

\subsection{Additional Experiment: Three-Class Dataset-Origin Classification}
As a diagnostic probe, we conducted an exploratory experiment that reframed the task from clinical classification to dataset-origin classification. The goal was not to improve screening performance, but to test whether the model's learned representation encodes dataset-specific characteristics strongly enough to separate images by their source. This question is motivated by a well-documented phenomenon in computer vision and medical imaging: models can often identify which dataset or institution an image originates from, exploiting acquisition-specific signatures rather than clinically meaningful content \cite{Liu2025-datasetbias}.

We kept the entire experimental setup unchanged, including the frozen Mammo-CLIP encoder, preprocessing, per-image normalization, and augmentation pipeline, and modified only the labeling objective. Rather than normal-versus-abnormal classification, each exam was labeled by its dataset of origin, yielding a three-class problem: NLBSD, CBIS-DDSM, or CMMD. Because the pipeline, normalization, and encoder are identical to the binary experiments, any ability to separate the three classes reflects residual dataset-specific structure that survives preprocessing rather than differences introduced by it. To ensure that separation could not be driven simply by unequal dataset sizes, we controlled for class imbalance using balanced batch sampling and report balanced (per-class) accuracy rather than raw accuracy.

This experiment is a diagnostic analysis, not a proposed model. Its purpose is to reveal whether screening data and externally sourced datasets occupy separable regions of the learned representation when the distinction is made explicit, thereby helping to explain the negative-transfer results observed above.

\paragraph{Controls and evaluation}
Every dataset passed through the same preprocessing pipeline, so that differences in outcome could not be attributed to how the images were prepared. Each image underwent identical DICOM intensity handling and per-image normalization, breast segmentation and cropping, and laterality alignment; augmentations were applied to the training split alone. Splits were always made at the patient level, ensuring that no patient's views appeared in more than one split. Every model used the same architecture and the same training hyperparameters, leaving the training data as the only variable that changed between experiments. One consequence of this design is that the test-set class balance differs between the NLBSD-only and mixed-domain settings, so prevalence-sensitive measures such as precision and F1 cannot be compared directly across them. We therefore treat AUC-ROC as the primary metric, reporting sensitivity and specificity alongside it as more clinically interpretable secondary measures.

\subsection{Statistical Analysis}
Discrimination was quantified by the area under the receiver operating characteristic curve (AUC), computed at the image level, with each mammographic view treated as one prediction; case-level results, obtained by max-pooling the views within an exam, are reported as a supplementary analysis. Ninety-five percent confidence intervals for each AUC were estimated by stratified bootstrap resampling (2{,}000 replicates, resampling positive and negative cases separately) and cross-checked against the DeLong interval~\cite{DeLong1988-mo}. Differences in AUC between the baseline and each mixed-training configuration were assessed on the shared NLBSD test images using DeLong's test for two correlated ROC curves, and the three $p$-values from these comparisons against the baseline were adjusted with the Holm--Bonferroni procedure~\cite{Holm1979-multiple}. For the operating-point metrics, a single decision threshold was selected on the baseline model and then applied unchanged to every model, so that sensitivity, specificity, and accuracy are compared at a common threshold rather than at a separately optimized point for each configuration; bootstrap 95\% confidence intervals are reported for each. Because test-set prevalence differed across evaluations, AUC served as the primary metric, and prevalence-dependent measures such as precision and F1-score were not used for cross-configuration comparison.

\section{Results}

\subsection{Baseline Performance on NLBSD}
Training and evaluating the model solely on the NLBSD screening cohort produced the highest overall performance, with an image-level AUC-ROC of 0.737 (95\% CI [0.686, 0.785]; Table~\ref{tab:results}). At the fixed operating point selected on this baseline, the model achieved an accuracy of 0.66, a sensitivity of 0.66, and a specificity of 0.66. This configuration serves as the primary reference because it reflects the exact screening distribution for which the system is intended.

Using this baseline, we address two questions in turn: whether adding external positives helps or harms performance on a true screening test set, and whether models trained on mixed datasets generalize across domains.

\begin{table}[H]
\centering
\caption{Comparison of screening performance under different training and testing combinations of NLBSD, CBIS-DDSM, and CMMD. The top block evaluates true screening performance on the fixed NLBSD test set; the bottom block evaluates mixed-domain test sets. Adding external abnormal-enriched datasets does not improve performance on the NLBSD screening distribution, and provides no meaningful gain under domain-mixed evaluation either.}
\begin{tabular}{l l c c c c}
\hline
\textbf{Train Set} & \textbf{Test Set} & \textbf{AUC-ROC} & \textbf{Acc} & \textbf{Sen} & \textbf{Spc} \\
\hline
NLBSD & NLBSD & \textbf{0.737} & \textbf{0.66} & \textbf{0.66} & \textbf{0.66} \\
NLBSD + CMMD$^{+}$ & NLBSD & 0.644 & 0.63 & 0.55 & 0.63 \\
NLBSD + DDSM$^{+}$ & NLBSD & 0.640 & 0.65 & 0.55 & 0.66 \\
NLBSD + DDSM$^{+}$ + CMMD$^{+}$ & NLBSD & 0.620 & 0.65 & 0.53 & 0.66 \\
\hline
\end{tabular}
\label{tab:results}
\end{table}

\subsection{Effect of Adding External Abnormal-Only Data}
The NLBSD-only model achieved the highest discrimination (image-level AUC $= 0.737$, 95\% CI $[0.686, 0.785]$). Adding external abnormal-enriched cases reduced performance in every configuration. Training on NLBSD\,+\,CMMD$^{+}$ lowered the AUC to $0.644$ ($[0.582, 0.703]$; $\Delta$AUC $= -0.093$), NLBSD\,+\,DDSM$^{+}$ to $0.640$ ($[0.576, 0.695]$; $\Delta$AUC $= -0.097$), and combining both external sources gave the lowest performance, $0.620$ ($[0.550, 0.684]$; $\Delta$AUC $= -0.117$). The degradation was monotonic, growing as more external sources were added.

All three reductions relative to the baseline were statistically significant (DeLong test for two correlated ROC curves, Holm-corrected $p = 0.015$, $0.005$, and $<0.001$ for the CMMD, DDSM, and combined configurations, respectively). We note that because each exam contributes four correlated views, the image-level test treats non-independent predictions as independent and its $p$-values should therefore be read as somewhat optimistic; the consistent, monotonic ordering across configurations, however, does not depend on this assumption.

Taken together, incorporating external abnormal-enriched datasets did not improve performance on the screening cohort. In every case it reduced the model's ability to discriminate on NLBSD, and the more external data were added, the larger the loss.
\begin{figure}[H]
\centering
\includegraphics[width=0.85\textwidth]{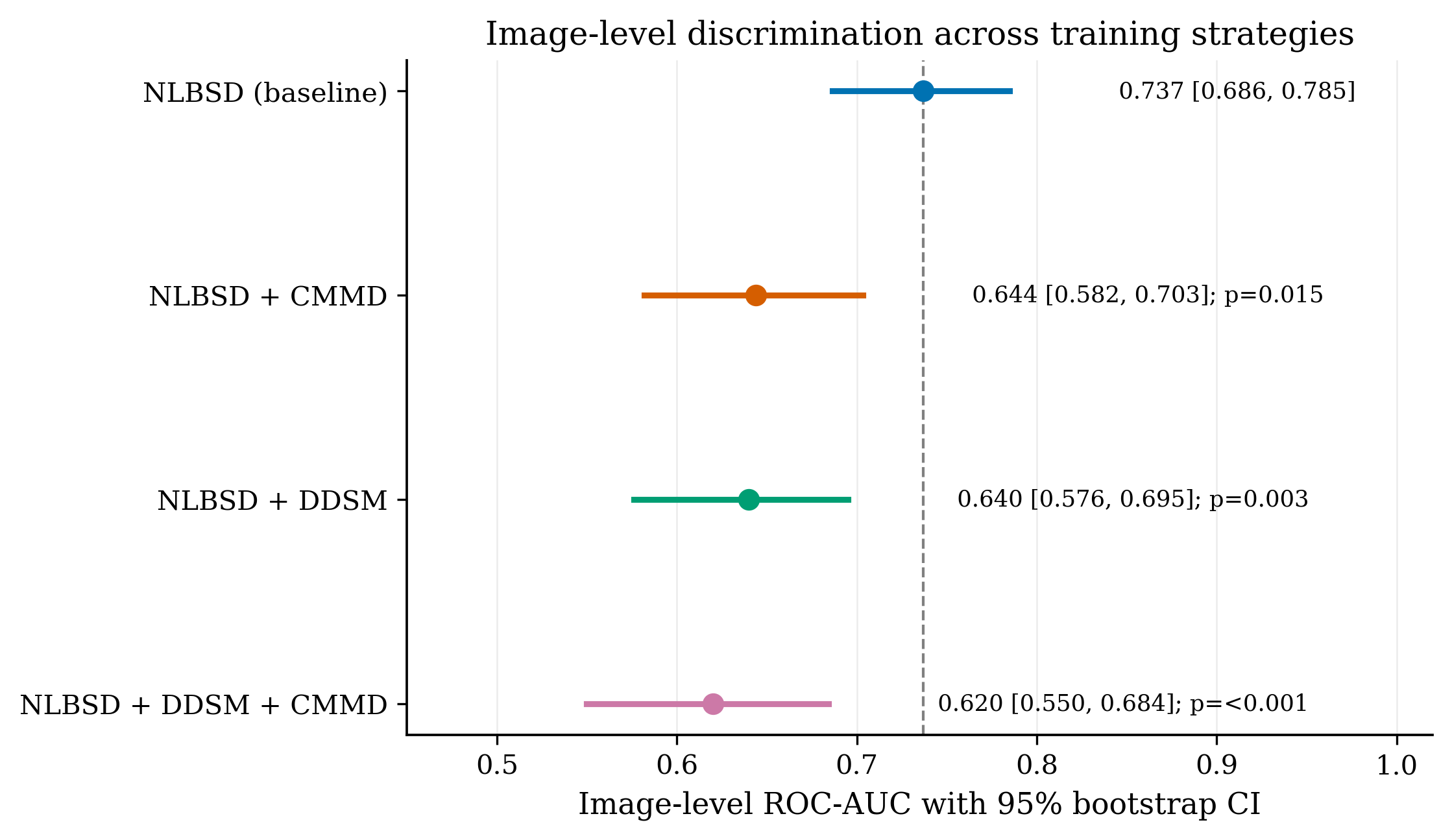}
\caption{Image-level AUC with 95\% bootstrap confidence intervals for each training configuration. Every mixed configuration falls below the NLBSD-only baseline (dashed line), with the reduction increasing monotonically as external sources are added. Differences relative to the baseline were significant by the paired DeLong test after Holm correction.}
\label{fig:forest}
\end{figure}

\subsection{Evaluation on Mixed-Domain Test Sets}
As a supplementary check, we evaluated each training configuration on its corresponding mixed-domain test set, where both the positive training and test cases were drawn partly from the external datasets. Performance was broadly similar to the Phase A results and did not exceed the NLBSD-only baseline: NLBSD\,+\,CMMD$^{+}$ reached an AUC of 0.653 on NLBSD\,$\cup$\,CMMD, NLBSD\,+\,CBIS-DDSM$^{+}$ reached 0.651 on NLBSD\,$\cup$\,CBIS-DDSM, and the fully combined configuration was lowest at 0.648. None of these settings surpassed the screening baseline, and we treat them only as a secondary observation rather than evidence of benefit from external data.

\subsection{Three-Class Dataset-Origin Results}
\begin{figure}[H]
\centering
\includegraphics[width=0.7\linewidth]{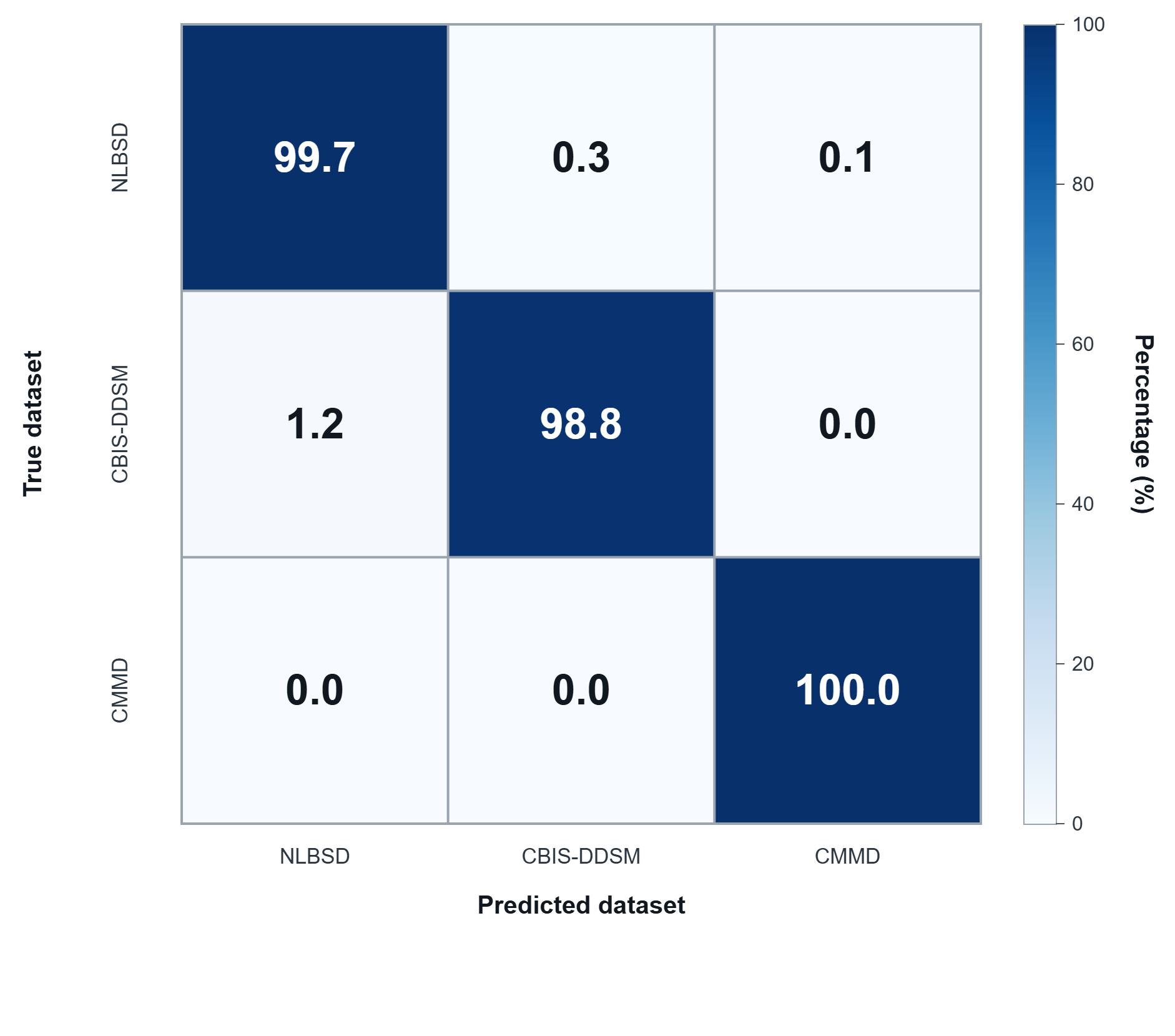}
\caption{Normalized confusion matrix for the three-class dataset-origin experiment. Rows correspond to true dataset origin and columns to predicted dataset; values are row-normalized percentages. The near-perfect diagonal (99.7\%, 98.8\%, 100.0\%; macro-average AUC 0.9998) shows that images from NLBSD, CBIS-DDSM, and CMMD are almost entirely separable by dataset-specific characteristics, despite identical preprocessing.}
\label{fig:confusion_three_class}
\end{figure}

When the task was reframed to predict dataset origin, the model separated the three datasets almost perfectly (Figure~\ref{fig:confusion_three_class}), with all classes reaching F1 above 0.98 and a macro-average AUC of 0.9998. Images from NLBSD, CBIS-DDSM, and CMMD therefore occupy highly separable regions of the learned feature space despite identical preprocessing and normalization, indicating that dataset-specific structure is strongly encoded even within a shared, mammography-pretrained representation.

\subsection{Summary of Findings}
Across all experiments, the strongest and most reliable performance came from training and evaluating the model exclusively on the NLBSD screening cohort. This configuration most closely reflects the intended clinical setting and consistently produced the highest AUC-ROC.

Introducing abnormal-enriched external datasets during training did not improve performance on the NLBSD screening test set. Performance declined relative to the NLBSD-only baseline in every configuration, even under identical preprocessing, normalization, and training procedures, and the decline grew as more external sources were added. Evaluations on mixed-domain test sets told the same story: no configuration surpassed the NLBSD-only model on true screening data.

The three-class experiment showed why. Images from NLBSD, CBIS-DDSM, and CMMD were almost perfectly separable by dataset origin, indicating that dataset-specific characteristics are strongly encoded in the learned representation even under a shared preprocessing pipeline.

Taken together, these results show that naïvely combining a screening cohort with abnormal-enriched external datasets provides no measurable benefit for screening performance. Instead, it introduces a dataset-dependent structure that the model can exploit, limiting generalization to the target screening population.

\section{Discussion}
This study set out to test a seemingly straightforward expectation: that adding biopsy-confirmed positive cases from external datasets would improve model performance on a specific screening cohort. Instead, we observed the opposite. The highest AUC-ROC on the NLBSD screening test set was obtained when both training and evaluation were restricted to NLBSD alone (AUC-ROC = 0.737), and performance degraded consistently when abnormal-enriched external datasets were introduced during training (AUC-ROC = 0.644 with a single external source and 0.620 when both CBIS-DDSM and CMMD positives were added). These results held even when evaluation was broadened to mixed-domain test sets that included external positive cases. Under every configuration considered here, naïvely mixing abnormal-only datasets failed to improve screening performance and, in most cases, reduced it.

\subsection{Why ``more data'' did not improve performance}
At first glance, the decline seems to contradict the intuition that more positive cases should strengthen discrimination. That intuition, however, holds only when the added samples come from the same or a closely related distribution as the target. Here that condition is explicitly violated. NLBSD is a population-level screening cohort with low prevalence, subtle findings, and a case mix dominated by normal exams, whereas CBIS-DDSM and CMMD are pathology-confirmed, abnormal-enriched collections with no normal screening negatives and a high proportion of clearly visible lesions. The added cases are therefore not simply ``more positives'' but positives drawn from a different domain.

Identical preprocessing does not remove this gap. Per-image normalization and cropping align surface formatting, but they cannot erase differences rooted in acquisition physics and dataset construction: film-derived images such as CBIS-DDSM retain digitization artifacts and grain structure, while digital datasets such as CMMD carry vendor-specific windowing and detector characteristics that produce distinct contrast and noise profiles. Population-level differences in breast density, age distribution, and imaging practice further shape tissue appearance. These signatures survive normalization, as our three-class experiment demonstrates directly: a model could identify an image's source dataset almost perfectly (macro-average AUC $0.9998$) despite the shared pipeline.

This is the same phenomenon documented across imaging domains. Classifiers reliably identify an image's source dataset in natural images \cite{Liu2025-datasetbias}, and a chest-radiograph model was shown to recognize its source hospital almost perfectly, exploiting acquisition signatures rather than pathology and generalizing poorly as a result \cite{Zech2018-cb}. In our setting, when abnormal-enriched sources are pooled with NLBSD, these source-specific features give the model an easier signal to fit than the subtle, low-prevalence patterns that define the screening task. The classifier is drawn toward decision boundaries that reflect the combined datasets' structure rather than the NLBSD distribution, and the added positives shift it toward detecting image source rather than underlying pathology. This negative transfer from domain shift is large enough to outweigh any benefit from the additional positive examples.

\subsection{Domain shift and negative transfer as primary drivers}
Taken together, the results point to domain shift and negative transfer as the main drivers of the decline. The degradation on NLBSD follows a clear pattern: models trained solely on NLBSD perform best, adding one external dataset reduces performance, and adding both sources produces the largest drop. This monotonic trend is difficult to attribute to random variation or unstable training. Evaluating on mixed-domain test sets does not reverse it: even when the test cohort includes external positives, performance stays below the NLBSD-only baseline. If the problem were simply the number of positive samples or the choice of threshold, at least one mixed configuration would be expected to match or surpass the baseline, and none did.

Beyond the reduction in AUC, the mixed-training models showed a clinically important shift in operating characteristics. Applying a single fixed decision threshold across all models, the NLBSD-only baseline balanced sensitivity and specificity (both $\approx 0.66$), whereas the mixed-training models retained similar specificity but lost sensitivity, falling to $0.53$--$0.55$. In a screening context this direction of change is precisely the wrong one: it corresponds to roughly $45$--$47\%$ of suspicious cases being missed, moving the model away from the high-sensitivity regime that screening triage requires. Naïvely adding abnormal-enriched external data therefore not only lowers overall discrimination but also biases the operating point away from the sensitivity-first setting appropriate for screening.

\subsection{Implications for dataset design and screening AI}
From a practical standpoint, these findings suggest that, at least for the model and training configuration studied here, screening performance is not improved by simply aggregating biopsy-confirmed cancer cases from external sources that do not reflect the target screening paradigm. Abnormal-enriched datasets derived from diagnostic or pathology workflows are not directly interchangeable with a screening cohort when used as additional positive samples for training a screening-triage model.

Successful multi-dataset training in this setting will more likely require explicit domain-aware strategies rather than naïve pooling. Candidate approaches include harmonizing intensity distributions and pixel-value mappings across scanners, site- or scanner-conditioned models, targeted augmentation that simulates cross-domain acquisition variation \cite{Zhang2020-mh}, and domain-adaptation methods that constrain the representation learned from external data to align with the target screening domain. Evaluating whether any of these can convert abnormal-enriched external data from a liability into a benefit is a natural direction for future work; our results establish the baseline problem that such methods would need to solve. Without such interventions, naïvely pooling heterogeneous datasets risks driving the model toward source-specific patterns that do not reflect real screening conditions.

\subsection{Limitations and future directions}
Several limitations should be acknowledged. First, all experiments used a single feature extractor, an EfficientNet-B5 encoder with Mammo-CLIP pretraining, applied as a frozen linear probe with a cross-entropy objective. Fine-tuning the encoder, using multi-view models, or adopting architectures and objectives designed for imbalanced or cross-domain settings may yield different behavior and could improve robustness. That said, the frozen-encoder design also strengthens our central observation: because the encoder was not adapted to any of the datasets, the near-perfect dataset separability we observe is baked into a general, mammography-pretrained representation rather than being an artifact of task-specific fine-tuning.

Second, although NLBSD provides a realistic screening cohort, it reflects the population, equipment, and acquisition protocol of a single regional program. Validation on additional true screening datasets would help establish how broadly these findings generalize.

Finally, our three-class experiment showed that dataset origin is strongly encoded in the learned representation even under a shared preprocessing pipeline. This was a diagnostic analysis, not a screening model, but it points to a clear need for learning strategies that explicitly account for domain differences. Future work should investigate domain-aware approaches, such as domain-adversarial training, representation alignment, or site-conditioned modeling, that reduce sensitivity to dataset origin while preserving clinically relevant features. Our results should therefore be read as an initial characterization of the challenges of mixing screening and cancer-enriched datasets, not as a strict limit on what more advanced methods or larger multi-institutional datasets might achieve.

\section{Conclusion}
Despite these limitations, this study offers a clear and cautionary finding: when model architecture, optimization, and evaluation are held constant, naïvely combining abnormal-enriched datasets with a true screening cohort does not improve, and often reduces, screening performance on a dataset such as NLBSD. Our results further show that dataset-specific characteristics are strongly reflected in the learned representation, indicating that models may implicitly separate data by source rather than learning a unified, screening-relevant concept of abnormality.

Together, these findings emphasize that, in mammography screening, the domain alignment of additional data matters as much as its quantity. Progress toward clinically reliable AI will depend on domain-aware strategies for integrating heterogeneous datasets, rather than unstructured pooling of all available cases. More broadly, our results help explain why the many available mammography datasets remain effectively siloed and why treating them as a single combined resource requires methods that explicitly bridge acquisition domains.

\section*{Compliance with Ethical Standards}
This retrospective study used open-access, de-identified human data from the Newfoundland and Labrador Breast Screening Program (NLBSD), available through the Federated Research Data Repository (FRDR). As the data are publicly available and fully de-identified, this study did not require additional institutional ethical approval.

\section*{Conflict of Interest}
This work was supported by the Seed, Bridge, and Multidisciplinary Fund at Memorial University of Newfoundland. The authors have no relevant financial or non-financial interests to disclose.

\section*{CRediT authorship contribution statement}
\textbf{Parham Hajishafiezahramini:} Conceptualization, Methodology, Software, Formal analysis, Investigation, Data curation, Visualization, Writing -- original draft.\\ \textbf{Matthew Hamilton:} Conceptualization, Methodology, Supervision, Writing -- review \& editing. \\ \textbf{Oscar Meruvia Pastor:} Supervision, Writing -- review \& editing. \\ \textbf{Edward Kendall:} Conceptualization, Resources, Data curation, Validation, Supervision, Writing -- review \& editing.
\section*{Data availability}
The Newfoundland and Labrador Breast Screening (NLBSD) dataset is publicly available through the Federated Research Data Repository (FRDR). The CBIS-DDSM and CMMD datasets are publicly available through The Cancer Imaging Archive (TCIA).

\bibliographystyle{elsarticle-num}
\bibliography{papers}

@ARTICLE{Sung2021-hq,
  title     = "Global cancer statistics 2020: {GLOBOCAN} estimates of incidence
               and mortality worldwide for 36 cancers in 185 countries",
  author    = "Sung, Hyuna and Ferlay, Jacques and Siegel, Rebecca L and
               Laversanne, Mathieu and Soerjomataram, Isabelle and Jemal,
               Ahmedin and Bray, Freddie",
  journal   = "CA Cancer J. Clin.",
  publisher = "Wiley",
  volume    =  71,
  number    =  3,
  pages     = "209--249",
  month     =  may,
  year      =  2021,
  copyright = "http://onlinelibrary.wiley.com/termsAndConditions\#vor",
  language  = "en"
}

@ARTICLE{Guan2022-bp,
  title     = "Domain adaptation for medical image analysis: A survey",
  author    = "Guan, Hao and Liu, Mingxia",

  journal   = "IEEE Trans. Biomed. Eng.",
  publisher = "Institute of Electrical and Electronics Engineers (IEEE)",
  volume    =  69,
  number    =  3,
  pages     = "1173--1185",
  month     =  mar,
  year      =  2022,
  copyright = "https://ieeexplore.ieee.org/Xplorehelp/downloads/license-information/IEEE.html",
  language  = "en"
}

@ARTICLE{Plunkett2022-us,
  title     = "Regional anesthesia for breast cancer surgery: which block is
               best? A review of the current literature",
  author    = "Plunkett, Anthony and Scott, Trevor L and Tracy, Erin",

  journal   = "Pain Manag.",
  publisher = "Informa UK Limited",
  volume    =  12,
  number    =  8,
  pages     = "943--950",
  month     =  nov,
  year      =  2022,
  language  = "en"
}

@ARTICLE{Parada2019-ww,
  title     = "Lifestyle patterns and survival following breast cancer in the
               Carolina breast cancer study",
  author    = "Parada, Jr, Humberto and Sun, Xuezheng and Tse, Chiu-Kit and
               Olshan, Andrew F and Troester, Melissa A",
  journal   = "Epidemiology",
  publisher = "Ovid Technologies (Wolters Kluwer Health)",
  volume    =  30,
  number    =  1,
  pages     = "83--92",
  month     =  jan,
  year      =  2019,
  language  = "en"
}

@misc{ccs2023breastcancer,
  author = {{Canadian Cancer Society}},
  title  = {Survival Statistics for Breast Cancer},
  year   = {2023},
  howpublished = {\url{https://cancer.ca/en/cancer-information/cancer-types/breast/prognosis-and-survival/survival-statistics}},
  note   = {Accessed: 2023-01-27}
}

@ARTICLE{Tabar2022-pv,
  title     = "A new approach to breast cancer terminology based on the
               anatomic site of tumour origin: The importance of radiologic
               imaging biomarkers",
  author    = "Tab{\'a}r, L{\'a}szl{\'o} and Dean, Peter B and Lee Tucker, F
               and Yen, Amy Ming-Fang and Chen, Sam Li-Sheng and Jen, Grace
               Hsiao Hsuan and Wang, Jackson Wei-Chun and Smith, Robert A and
               Duffy, Stephen W and Chen, Tony Hsiu-Hsi",

  journal   = "Eur. J. Radiol.",
  publisher = "Elsevier BV",
  volume    =  149,
  number    =  110189,
  pages     = "110189",
  month     =  apr,
  year      =  2022,

             
  language  = "en"
}

@ARTICLE{Tabar2019-wr,
  title     = "The incidence of fatal breast cancer measures the increased
               effectiveness of therapy in women participating in mammography
               screening",
  author    = "Tab{\'a}r, L{\'a}szl{\'o} and Dean, Peter B and Chen, Tony
               Hsiu-Hsi and Yen, Amy Ming-Fang and Chen, Sam Li-Sheng and Fann,
               Jean Ching-Yuan and Chiu, Sherry Yueh-Hsia and Ku, May Mei-Sheng
               and Wu, Wendy Yi-Ying and Hsu, Chen-Yang and Chen, Yu-Ching and
               Beckmann, Kerri and Smith, Robert A and Duffy, Stephen W",

  journal   = "Cancer",
  publisher = "Wiley",
  volume    =  125,
  number    =  4,
  pages     = "515--523",
  month     =  feb,
  year      =  2019,
  copyright = "http://creativecommons.org/licenses/by-nc-nd/4.0/",
  language  = "en"
}

@ARTICLE{Duffy2020-we,
  title     = "Mammography screening reduces rates of advanced and fatal breast
               cancers: Results in 549,091 women",
  author    = "Duffy, Stephen W and Tab{\'a}r, L{\'a}szl{\'o} and Yen, Amy
               Ming-Fang and Dean, Peter B and Smith, Robert A and Jonsson,
               H{\aa}kan and T{\"o}rnberg, Sven and Chen, Sam Li-Sheng and
               Chiu, Sherry Yueh-Hsia and Fann, Jean Ching-Yuan and Ku, May
               Mei-Sheng and Wu, Wendy Yi-Ying and Hsu, Chen-Yang and Chen,
               Yu-Ching and Svane, Gunilla and Azavedo, Edward and
               Grundstr{\"o}m, Helene and Sund{\'e}n, Per and Leifland, Karin
               and Frodis, Ewa and Ramos, Joakim and Epstein, Birgitta and
               {\AA}kerlund, Anders and Sundbom, Ann and Bord{\'a}s, P{\'a}l
               and Wallin, Hans and Starck, Leena and Bj{\"o}rkgren, Annika and
               Carlson, Stina and Fredriksson, Irma and Ahlgren, Johan and
               {\"O}hman, Daniel and Holmberg, Lars and Chen, Tony Hsiu-Hsi",
 
  journal   = "Cancer",
  publisher = "Wiley",
  volume    =  126,
  number    =  13,
  pages     = "2971--2979",
  month     =  jul,
  year      =  2020,
  copyright = "http://creativecommons.org/licenses/by-nc-nd/4.0/",
  language  = "en"
}

@ARTICLE{Sweeney2017-ow,
 title     = "A review of mammographic positioning image quality criteria for
               the craniocaudal projection",
  author    = "Sweeney, Rhonda-Joy I and Lewis, Sarah J and Hogg, Peter and
               McEntee, Mark F",
  journal   = "Br. J. Radiol.",
  publisher = "British Institute of Radiology",
  volume    =  91,
  number    =  1082,
  pages     = "20170611",
  month     =  feb,
  year      =  2018,
  language  = "en"
}

@ARTICLE{Sardanelli2017-ua,
  title     = "Position paper on screening for breast cancer by the European
               society of breast imaging ({EUSOBI}) and 30 national breast
               radiology bodies from Austria, Belgium, Bosnia and Herzegovina,
               Bulgaria, Croatia, Czech Republic, Denmark, Estonia, Finland,
               France, Germany, Greece, Hungary, Iceland, Ireland, Italy,
               Israel, Lithuania, Moldova, the Netherlands, Norway, Poland,
               Portugal, Romania, Serbia, Slovakia, Spain, Sweden, Switzerland
               and turkey",
  author    = "Sardanelli, Francesco and Aase, Hildegunn S and {\'A}lvarez,
               Marina and Azavedo, Edward and Baarslag, Henk J and Balleyguier,
               Corinne and Baltzer, Pascal A and Beslagic, Vanesa and Bick,
               Ulrich and Bogdanovic-Stojanovic, Dragana and Briediene, Ruta
               and Brkljacic, Boris and Camps Herrero, Julia and Colin,
               Catherine and Cornford, Eleanor and Danes, Jan and de Geer,
               G{\'e}rard and Esen, Gul and Evans, Andrew and Fuchsjaeger,
               Michael H and Gilbert, Fiona J and Graf, Oswald and Hargaden,
               Gormlaith and Helbich, Thomas H and Heywang-K{\"o}brunner,
               Sylvia H and Ivanov, Valentin and J{\'o}nsson, {\'A}sbj{\"o}rn
               and Kuhl, Christiane K and Lisencu, Eugenia C and Luczynska,
               Elzbieta and Mann, Ritse M and Marques, Jose C and Martincich,
               Laura and Mortier, Margarete and M{\"u}ller-Schimpfle, Markus
               and Ormandi, Katalin and Panizza, Pietro and Pediconi, Federica
               and Pijnappel, Ruud M and Pinker, Katja and Rissanen, Tarja and
               Rotaru, Natalia and Saguatti, Gianni and Sella, Tamar and
               Slobodn{\'\i}kov{\'a}, Jana and Talk, Maret and Taourel, Patrice
               and Trimboli, Rubina M and Vejborg, Ilse and Vourtsis, Athina
               and Forrai, Gabor",
  journal   = "Eur. Radiol.",
  publisher = "Springer Nature",
  volume    =  27,
  number    =  7,
  pages     = "2737--2743",
  month     =  jul,
  year      =  2017
}

@misc{acs_mammograms,
  author = {{American Cancer Society}},
  title  = {Limitations of Mammograms},
  year   = {2023},
  url    = {https://www.cancer.org/cancer/types/breast-cancer/screening-tests-and-early-detection/mammograms/limitations-of-mammograms.html},
  note   = {Accessed: 2024-10-18}
}

@ARTICLE{Kendall2013-ec,
  title     = "Automatic detection of anomalies in screening mammograms",
  author    = "Kendall, Edward J and Barnett, Michael G and Chytyk-Praznik,
               Krista",
  journal   = "BMC Med. Imaging",
  publisher = "Springer Science and Business Media LLC",
  volume    =  13,
  number    =  1,
  pages     = "43",
  month     =  dec,
  year      =  2013,
  language  = "en"
}

@ARTICLE{Kendall2014-zg,
  title     = "Automated breast image classification using features from its
               discrete cosine transform",
  author    = "Kendall, Edward J and Flynn, Matthew T",
  journal   = "PLoS One",
  publisher = "Public Library of Science (PLoS)",
  volume    =  9,
  number    =  3,
  pages     = "e91015",
  month     =  mar,
  year      =  2014,
  language  = "en"
}

@ARTICLE{Halling-Brown2021-yp,
  title     = "{OPTIMAM} mammography image database: A large-scale resource of
               mammography images and clinical data",
  author    = "Halling-Brown, Mark D and Warren, Lucy M and Ward, Dominic and
               Lewis, Emma and Mackenzie, Alistair and Wallis, Matthew G and
               Wilkinson, Louise S and Given-Wilson, Rosalind M and McAvinchey,
               Rita and Young, Kenneth C",
  journal   = "Radiol. Artif. Intell.",
  publisher = "Radiological Society of North America (RSNA)",
  volume    =  3,
  number    =  1,
  pages     = "e200103",
  month     =  jan,
  year      =  2021,
  language  = "en"
}

@inproceedings{heath2001digital,
  author    = {M. Heath and others},
  title     = {Current Status of the Digital Database for Screening Mammography},
  booktitle = {Proceedings of the Fifth International Workshop on Digital Mammography},
  pages     = {212--218},
  year      = {2001},
  editor    = {M. J. Yaffe},
  publisher = {Medical Physics Publishing}
}

@ARTICLE{Lee2017-fa,
  title     = "A curated mammography data set for use in computer-aided
               detection and diagnosis research",
  author    = "Lee, Rebecca Sawyer and Gimenez, Francisco and Hoogi, Assaf and
               Miyake, Kanae Kawai and Gorovoy, Mia and Rubin, Daniel L",
  journal   = "Sci. Data",
  publisher = "Springer Science and Business Media LLC",
  volume    =  4,
  number    =  1,
  pages     = "170177",
  month     =  dec,
  year      =  2017,
  copyright = "https://creativecommons.org/licenses/by/4.0",
  language  = "en"
}

@MISC{Suckling2015-al,
title     = "Mammographic Image Analysis Society ({MIAS}) database v1.21",
  author    = "Suckling, John and Parker, J and Dance, D and Astley, S and
               Hutt, I and Boggis, C and Ricketts, I and Stamatakis, E and
               Cerneaz, N and Kok, S and Taylor, P and Betal, D and Savage, J",
  publisher = "Apollo - University of Cambridge Repository",
  year      =  2015
}

@ARTICLE{Moreira2012-si,
  title     = "{INbreast}: toward a full-field digital mammographic database",
  author    = "Moreira, In{\^e}s C and Amaral, Igor and Domingues, In{\^e}s and
               Cardoso, Ant{\'o}nio and Cardoso, Maria Jo{\~a}o and Cardoso,
               Jaime S",
  journal   = "Acad. Radiol.",
  publisher = "Elsevier BV",
  volume    =  19,
  number    =  2,
  pages     = "236--248",
  month     =  feb,
  year      =  2012,
  language  = "en"
}

@ARTICLE{Nguyen2023-zg,
  title    = "{VinDr-Mammo}: A large-scale benchmark dataset for computer-aided
              diagnosis in full-field digital mammography",
  author   = "Nguyen, Hieu T and Nguyen, Ha Q and Pham, Hieu H and Lam, Khanh
              and Le, Linh T and Dao, Minh and Vu, Van",
  journal  = "Sci. Data",
  volume   =  10,
  number   =  1,
  pages    = "277",
  month    =  may,
  year     =  2023,
  language = "en"
}

@ARTICLE{Cai2023-wk,
  title    = "An online mammography database with biopsy confirmed types",
  author   = "Cai, Hongmin and Wang, Jinhua and Dan, Tingting and Li, Jiao and
              Fan, Zhihao and Yi, Weiting and Cui, Chunyan and Jiang, Xinhua
              and Li, Li",
  journal  = "Sci. Data",
  volume   =  10,
  number   =  1,
  pages    = "123",
  month    =  mar,
  year     =  2023,
  language = "en"
}

@misc{carr2022rsna,
  author = {C. Carr and others},
  title = {RSNA Screening Mammography Breast Cancer Detection},
  year = {2022},
  howpublished = {\url{https://kaggle.com/competitions/rsna-breast-cancer-detection}},
  note = {Accessed: 2023-01-10}
}

@ARTICLE{Jeong2023-cl,
  title     = "The {EMory} {BrEast} imaging dataset ({EMBED)}: A racially
               diverse, granular dataset of 3.4 million screening and
               diagnostic mammographic images",
  author    = "Jeong, Jiwoong J and Vey, Brianna L and Bhimireddy, Ananth and
               Kim, Thomas and Santos, Thiago and Correa, Ramon and Dutt, Raman
               and Mosunjac, Marina and Oprea-Ilies, Gabriela and Smith,
               Geoffrey and Woo, Minjae and McAdams, Christopher R and Newell,
               Mary S and Banerjee, Imon and Gichoya, Judy and Trivedi, Hari",
   journal   = "Radiol. Artif. Intell.",
  publisher = "Radiological Society of North America (RSNA)",
  volume    =  5,
  number    =  1,
  pages     = "e220047",
  month     =  jan,
  year      =  2023,
  language  = "en"
}

@ARTICLE{Frazer2023-zn,
  title    = "{ADMANI}: Annotated digital mammograms and associated non-image
              datasets",
  author   = "Frazer, Helen M L and Tang, Jennifer S N and Elliott, Michael S
              and Kunicki, Katrina M and Hill, Brendan and Karthik, Ravishankar
              and Kwok, Chun Fung and Pe{\~n}a-Solorzano, Carlos A and Chen,
              Yuanhong and Wang, Chong and Al-Qershi, Osamah and Fox, Samantha
              K and Li, Shuai and Makalic, Enes and Nguyen, Tuong L and
              Schmidt, Daniel F and Basnayake Ralalage, Prabhathi and Lippey,
              Jocelyn F and Brotchie, Peter and Hopper, John L and Carneiro,
              Gustavo and McCarthy, Davis J",
  journal  = "Radiol. Artif. Intell.",
  volume   =  5,
  number   =  2,
  pages    = "e220072",
  month    =  mar,
  year     =  2023,
  language = "en"
}

@ARTICLE{Jafari2023-zu,
  title     = "Breast cancer detection in mammography images: A {CNN-based}
               approach with feature selection",
  author    = "Jafari, Zahra and Karami, Ebrahim",
  journal   = "Information (Basel)",
  publisher = "MDPI AG",
  volume    =  14,
  number    =  7,
  pages     = "410",
  month     =  jul,
  year      =  2023,
  copyright = "https://creativecommons.org/licenses/by/4.0/",
  language  = "en"
}

@ARTICLE{Sahu2023-jj,
  title     = "High accuracy hybrid {CNN} classifiers for breast cancer
               detection using mammogram and ultrasound datasets",
  author    = "Sahu, Adyasha and Das, Pradeep Kumar and Meher, Sukadev",
  journal   = "Biomed. Signal Process. Control",
  publisher = "Elsevier BV",
  volume    =  80,
  number    =  104292,
  pages     = "104292",
  month     =  feb,
  year      =  2023,
  language  = "en"
}

@ARTICLE{Raaj2023-li,
  title     = "Breast cancer detection and diagnosis using hybrid deep learning
               architecture",
  author    = "Raaj, R Sathesh",
  journal   = "Biomed. Signal Process. Control",
  publisher = "Elsevier BV",
  volume    =  82,
  number    =  104558,
  pages     = "104558",
  month     =  apr,
  year      =  2023,
  language  = "en"
}

@ARTICLE{Arnold2022-vq,
  title     = "Current and future burden of breast cancer: Global statistics
               for 2020 and 2040",
  author    = "Arnold, Melina and Morgan, Eileen and Rumgay, Harriet and Mafra,
               Allini and Singh, Deependra and Laversanne, Mathieu and Vignat,
               Jerome and Gralow, Julie R and Cardoso, Fatima and Siesling,
               Sabine and Soerjomataram, Isabelle",

  journal   = "Breast",
  publisher = "Elsevier BV",
  volume    =  66,
  pages     = "15--23",
  month     =  dec,
  year      =  2022,
  copyright = "http://creativecommons.org/licenses/by-nc-nd/3.0/igo/",
  language  = "en"
}

@ARTICLE{Dembrower2020-ul,
  title     = "A multi-million mammography image dataset and population-based
               screening cohort for the training and evaluation of deep neural
               networks-the cohort of {Screen-Aged} Women ({CSAW})",
  author    = "Dembrower, Karin and Lindholm, Peter and Strand, Fredrik",
  journal   = "J. Digit. Imaging",
  publisher = "Springer Science and Business Media LLC",
  volume    =  33,
  number    =  2,
  pages     = "408--413",
  month     =  apr,
  year      =  2020,
  copyright = "https://creativecommons.org/licenses/by/4.0",
  language  = "en"
}

@ARTICLE{Giaquinto2024-yk,
  title     = "Breast cancer statistics 2024",
  author    = "Giaquinto, Angela N and Sung, Hyuna and Newman, Lisa A and
               Freedman, Rachel A and Smith, Robert A and Star, Jessica and
               Jemal, Ahmedin and Siegel, Rebecca L",

  journal   = "CA Cancer J. Clin.",
  publisher = "Wiley",
  volume    =  74,
  number    =  6,
  pages     = "477--495",
  month     =  nov,
  year      =  2024,
  copyright = "http://creativecommons.org/licenses/by-nc-nd/4.0/",
  language  = "en"
}

@ARTICLE{Abdikenov2025-jd,
  title     = "Innovative multi-view strategies for {AI-assisted} breast cancer
               detection in mammography",
  author    = "Abdikenov, Beibit and Zhaksylyk, Tomiris and Imasheva, Aruzhan
               and Orazayev, Yerzhan and Karibekov, Temirlan",
  journal   = "J. Imaging",
  publisher = "MDPI AG",
  volume    =  11,
  number    =  8,
  pages     = "247",
  month     =  jul,
  year      =  2025,
  copyright = "https://creativecommons.org/licenses/by/4.0/",
  language  = "en"
}

@ARTICLE{Kendall2025-mv,

  title     = "Full field digital mammography dataset from a population
               screening program",
  author    = "Kendall, Edward and Hajishafiezahramini, Parham and Hamilton,
               Matthew and Doyle, Gregory and Wadden, Nancy and Meruvia-Pastor,
               Oscar",
 
  journal   = "Sci. Data",
  publisher = "Springer Science and Business Media LLC",
  volume    =  12,
  number    =  1,
  pages     = "1479",
  month     =  aug,
  year      =  2025,
  copyright = "https://creativecommons.org/licenses/by-nc-nd/4.0",
  language  = "en"
}

@MISC{Kendall2025-jn,

  author    = {Kendall, Edward and Hajishafiezahramini, Parham and Hamilton, Matthew and Doyle, Gregory and Wadden, Nancy and Meruvia-Pastor, Oscar},
  title     = {Newfoundland and Labrador Breast Screening ({NLBS}) Dataset ({NL-Breast-Screen})},
  publisher = {Federated Research Data Repository},
  year      = {2025},
  doi       = {10.20383/103.01526},
  url       = {https://doi.org/10.20383/103.01526}
}

@ARTICLE{Taskiran2025-rt,
  title    = "A comprehensive evaluation of oversampling techniques for
              enhancing text classification performance",
  author   = "Taskiran, Salimkan Fatma and Turkoglu, Bahaeddin and Kaya, Ersin
              and Asuroglu, Tunc",
  journal  = "Sci. Rep.",
  volume   =  15,
  number   =  1,
  pages    = "21631",
  month    =  jul,
  year     =  2025,
  language = "en"
}

@ARTICLE{Velarde2024-cm,
  title    = "Robustness of deep networks for mammography: Replication across
              public datasets",
  author   = "Velarde, Osvaldo M and Lin, Clarissa and Eskreis-Winkler, Sarah
              and Parra, Lucas C",
  
  journal  = "J. Imaging Inform. Med.",
  volume   =  37,
  number   =  2,
  pages    = "536--546",
  month    =  apr,
  year     =  2024,
  language = "en"
}

@ARTICLE{Wang2020-ez,
  title     = "Inconsistent performance of deep learning models on mammogram
               classification",
  author    = "Wang, Xiaoqin and Liang, Gongbo and Zhang, Yu and Blanton,
               Hunter and Bessinger, Zachary and Jacobs, Nathan",

  journal   = "J. Am. Coll. Radiol.",
  publisher = "Elsevier BV",
  volume    =  17,
  number    =  6,
  pages     = "796--803",
  month     =  jun,
  year      =  2020,
  language  = "en"
}

@inproceedings{Liu2025-datasetbias,
  author    = {Liu, Zhuang and He, Kaiming},
  title     = {A Decade's Battle on Dataset Bias: Are We There Yet?},
  booktitle = {International Conference on Learning Representations (ICLR)},
  year      = {2025},
  note      = {arXiv:2403.08632},
}

@ARTICLE{DeLong1988-mo,
  title    = "Comparing the areas under two or more correlated receiver operating characteristic curves: a nonparametric approach",
  author   = "DeLong, E R and DeLong, D M and Clarke-Pearson, D L",

  journal  = "Biometrics",
  volume   =  44,
  number   =  3,
  pages    = "837--845",
  month    =  sep,
  year     =  1988,
  address  = "England",
  language = "en"
}

@ARTICLE{Zech2018-cb,
  title     = "Variable generalization performance of a deep learning model to
               detect pneumonia in chest radiographs: A cross-sectional study",
  author    = "Zech, John R and Badgeley, Marcus A and Liu, Manway and Costa,
               Anthony B and Titano, Joseph J and Oermann, Eric Karl",

  journal   = "PLoS Med.",
  publisher = "Public Library of Science (PLoS)",
  volume    =  15,
  number    =  11,
  pages     = "e1002683",
  month     =  nov,
  year      =  2018,
  copyright = "http://creativecommons.org/licenses/by/4.0/",
  language  = "en"
}

@ARTICLE{Zhang2020-mh,
  title     = "Generalizing deep learning for medical image segmentation to
               unseen domains via deep stacked transformation",
  author    = "Zhang, Ling and Wang, Xiaosong and Yang, Dong and Sanford,
               Thomas and Harmon, Stephanie and Turkbey, Baris and Wood,
               Bradford J and Roth, Holger and Myronenko, Andriy and Xu,
               Daguang and Xu, Ziyue",

  journal   = "IEEE Trans. Med. Imaging",
  publisher = "Institute of Electrical and Electronics Engineers (IEEE)",
  volume    =  39,
  number    =  7,
  pages     = "2531--2540",
  month     =  jul,
  year      =  2020,
  copyright = "https://ieeexplore.ieee.org/Xplorehelp/downloads/license-information/IEEE.html",
  language  = "en"
}

@article{Holm1979-multiple,
 ISSN = {03036898, 14679469},
 URL = {http://www.jstor.org/stable/4615733},
 author = {Sture Holm},
 journal = {Scandinavian Journal of Statistics},
 number = {2},
 pages = {65--70},
 publisher = {[Board of the Foundation of the Scandinavian Journal of Statistics, Wiley]},
 title = {A Simple Sequentially Rejective Multiple Test Procedure},
 year = {1979},
 volume = {6}
}

@INCOLLECTION{Ghosh2024-lb,
  title     = "{Mammo-CLIP}: A vision language foundation model to enhance data
               efficiency and robustness in mammography",
  booktitle = "Medical Image Computing and Computer Assisted Intervention -- MICCAI 2024",
  author    = "Ghosh, Shantanu and Poynton, Clare B and Visweswaran, Shyam and
               Batmanghelich, Kayhan",
  publisher = "Springer Nature Switzerland",
  pages     = "632--642",
 
  year      =  2024,
  address   = "Cham",
  copyright = "https://www.springernature.com/gp/researchers/text-and-data-mining",
  language  = "en"
}

@InProceedings{pmlr-v97-tan19a,
  title = 	 {{E}fficient{N}et: Rethinking Model Scaling for Convolutional Neural Networks},
  author =       {Tan, Mingxing and Le, Quoc},
  booktitle = 	 {Proceedings of the 36th International Conference on Machine Learning},
  pages = 	 {6105--6114},
  year = 	 {2019},
  editor = 	 {Chaudhuri, Kamalika and Salakhutdinov, Ruslan},
  volume = 	 {97},
  series = 	 {Proceedings of Machine Learning Research},
  month = 	 {09--15 Jun},
  publisher =    {PMLR},
  url = 	 {https://proceedings.mlr.press/v97/tan19a.html}
}

@ARTICLE{Loshchilov2017-un,
  title         = "Decoupled weight decay regularization",
  author        = "Loshchilov, Ilya and Hutter, Frank",
 
  month         =  nov,
  year          =  2017,
  copyright     = "http://arxiv.org/licenses/nonexclusive-distrib/1.0/",
  archivePrefix = "arXiv",
  primaryClass  = "cs.LG",
  eprint        = "1711.05101"
}

\end{document}